\documentclass[10pt,twocolumn,letterpaper]{article}

\PassOptionsToPackage{table}{xcolor}

\usepackage{cvpr}              % To produce the CAMERA-READY version
\definecolor{cvprblue}{rgb}{0.21,0.49,0.74}
\usepackage[pagebackref,breaklinks,colorlinks,allcolors=cvprblue]{hyperref}

\usepackage{multirow}
\usepackage{xcolor}
\newcommand{\cmark}{\ding{51}}
\newcommand{\xmark}{\ding{55}}
\usepackage{graphicx} 
\definecolor{SkyBlue}{RGB}{190, 230, 255}
\usepackage{wrapfig}
\usepackage{amsmath}
\usepackage{subcaption}
\usepackage{adjustbox}
\usepackage{pifont}

\def\paperID{4606} % *** Enter the Paper ID here
\def\confName{CVPR}
\def\confYear{2026}

\title{ESAM++: Efficient Online 3D Perception on the Edge}

\author{
Qin Liu$^1$~~~~
Lavisha Aggarwal$^2$~~~~
Saptarashmi Bandyopadhyay$^2$~~~~ 
Vikas Bahirwani$^2$~~~~ \\
Marc Niethammer$^3$~~~~
Ehsan Adeli$^1$~~~~
Andrea Colaco$^2$ \\
$^1$Stanford University~~~~
$^2$Google~~~~
$^3$UC San Diego \\
{\tt\small \href{https://github.com/qinliuliuqin/esamplusplus}{https://github.com/qinliuliuqin/esamplusplus}}
}

\begin{document}
\maketitle
\begin{abstract}
Online 3D scene perception in real time is essential for robotics, AR/VR, and autonomous systems, particularly in edge computing scenarios where computational resources are limited and privacy is crucial. Recent state-of-the-art methods like EmbodiedSAM (ESAM) demonstrate the promise of online 3D perception by leveraging the Segment Anything Model (SAM) for real-time, fine-grained, and generalized 3D instance segmentation. However, ESAM still relies on a computationally expensive 3D sparse UNet for point cloud feature extraction, which accounts for the majority of the 3D inference time, hindering its practicality on resource-constrained devices. In this paper, we propose \textbf{ESAM++}, a lightweight and scalable alternative for online 3D scene perception tailored to edge devices without GPU acceleration. Our method introduces a 3D Sparse Feature Pyramid Network (SFPN) that efficiently captures multi-scale geometric features from streaming 3D point clouds while significantly reducing computational overhead and model size. We evaluate our approach on four challenging segmentation benchmarks, namely ScanNet, ScanNet200, SceneNN, and 3RScan, demonstrating that our model achieves competitive accuracy with up to 3$\times$ faster inference with a 2$\times$ smaller model size compared to ESAM, enabling practical deployment on edge devices.
\end{abstract}    
\section{Introduction}
\label{sec:intro}

Real-time online 3D scene perception plays a central role in a wide range of applications, including robotics~\cite{mousavian20196,chaplot2020object,zhang20233d}, augmented and extended reality (AR/VR)~\cite{du2020depthlab,piekarski2002arquake}, and autonomous navigation~\cite{bojarski2016end,kendall2019learning,giusti2015machine}. In these settings, the ability to perform accurate and low-latency segmentation directly from streaming RGB-D videos is critical. This is especially important for applications requiring on-device inference, where computational resources are limited and data privacy is a primary concern. However, most existing 3D perception approaches~\cite{yang2023sam3d,xu2024memory,liu2022ins} are designed with server-grade hardware in mind and fall short of meeting the stringent requirements for efficiency, latency, and privacy.

\begin{figure*}
  \centering
  \includegraphics[width=\textwidth]{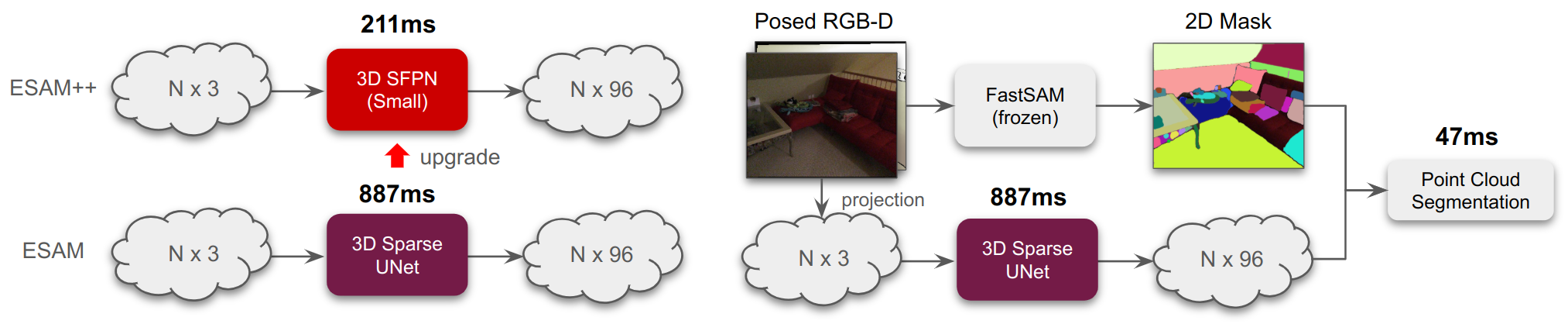}
  \caption{\textbf{Overview of ESAM and ESAM++}. ESAM~\cite{xu2024embodiedsam} (right figure) is the state-of-the-art approach for online 3D scene perception. We identify two key efficiency bottlenecks in its design: (1) the use of a frozen Visual Foundation Model (VFM) based on FastSAM~\cite{zhao2023fast}, and (2) a point cloud encoder built upon a 3D sparse UNet. This work focuses on optimizing the latter, while improvements to the VFM component are left for future research. The time cost is evaluated using an Intel Xeon Silver 4314 CPU@2.40GHz.}
  \label{fig:esam_framework_with_efficiency}
\end{figure*}

Recent work has made significant strides toward bridging the gap. Notably, EmbodiedSAM (ESAM)~\cite{xu2024embodiedsam} introduces an online 3D perception method that leverages the Segment Anything Model (SAM)~\cite{kirillov2023segment,zhao2023fast} for real-time, fine-grained, and generalized 3D instance segmentation. By encoding 2D segmentation masks as 3D geometric-aware queries, ESAM enables efficient mask merging across frames and achieves strong performance on multiple benchmarks. This design not only leverages the powerful generalization capabilities of SAM but also offers a promising way to integrate 2D and 3D information in an online setting. However, the framework still relies on a computationally expensive 3D sparse UNet~\cite{choy20194d} for point cloud feature extraction, which accounts for the majority of the 3D inference time, as shown in Figure~\ref{fig:esam_framework_with_efficiency}. This significant computational cost hinders its practicality on resource-constrained devices. Therefore, the \textbf{3D sparse UNet is identified as the primary efficiency bottleneck} in ESAM. Developing a more efficient alternative to the 3D sparse UNet for point cloud feature extraction remains an urgent and open challenge.

Despite its effectiveness in capturing spatial context, the 3D sparse UNet suffers from significant efficiency limitations for point cloud feature extraction. \textbf{\emph{First}}, the 3D sparse UNet produces the final output \emph{exclusively} at the last layer that operates on data with the same resolution as the input. As a result, when the input resolution is high or the number of output channels is large, these high-resolution layers incur substantial computational costs. This is confirmed by our per-layer computational analysis in Section~\ref{sec:method}. \textbf{\emph{Second}}, as the input resolution progressively decreases throughout the encoder, the number of channels typically increases, resulting in a larger number of model parameters, which can pose a bottleneck in terms of model size. This is also confirmed by our per-layer computational analysis in Section~\ref{sec:method}. \textbf{\emph{Finally}}, the decoder of the 3D sparse UNet progressively upsamples hierarchical features to the original resolution, yielding rich multi-scale representations. However, predictions are made only at the highest resolution, leaving lower-resolution features underutilized. This design choice limits efficiency and raises a key question: \emph{can multi-scale predictions be leveraged to better exploit these intermediate layers?} Architectures such as Feature Pyramid Networks (FPN)~\cite{lin2017feature}, which enable multi-scale predictions, are well-established in 2D computer vision but remain underexplored in the domain of online 3D scene perception.

To address these limitations, we propose ESAM++, a lightweight and scalable online 3D scene perception framework that incorporates a 3D sparse feature pyramid (SFPN) for efficient point cloud feature extraction. SFPN reduces computational cost by limiting the number of output channels in high-resolution layers. To compensate for this reduction, it fully leverages the decoder's feature hierarchy and employs efficient multi-scale feature aggregation. Each level of the pyramid is carefully optimized to balance spatial resolution and computational latency, making SFPN particularly well-suited for latency-sensitive applications. In addition, we further reduce the model size by removing intermediate layers that demand an excessive number of channels. 

We evaluate our method on four challenging 3D instance segmentation benchmarks—ScanNet, ScanNet200, SceneNN, and 3RScan—and demonstrate that our model achieves competitive performance while running up to 3$\times$ faster inference with 2$\times$ smaller model size compared to ESAM. Moreover, ESAM++ demonstrates strong potential in data-efficient settings, maintaining competitive performance even when trained on limited data. It also exhibits robustness to camera noise, highlighting its promise for RGB-only online 3D scene perception when integrated with RGB-only online reconstruction models such as CUT3R~\cite{wang2025continuous}. Our contributions are summarized as follows:

\begin{itemize}[leftmargin=*]
\item We analyze the computational inefficiencies of ESAM's 3D sparse UNet and identify it as the primary bottleneck in online 3D perception. This analysis serves as a foundation for designing a more efficient backbone for point cloud feature extraction.

\item We introduce a novel 3D sparse feature pyramid (SFPN) for efficient multi-scale point cloud representation, specifically designed for online 3D scene perception on edge devices without GPU acceleration.

\item We demonstrate that ESAM++ delivers up to 3$\times$ faster inference and 2$\times$ smaller model size than ESAM, while maintaining competitive accuracy across challenging benchmarks. It also excels in low-data regimes and noisy conditions, enabling future extension for RGB-only online 3D scene perception.
\end{itemize}
\section{Related Work}

\paragraph{Efficient Point Cloud Representation.}
PointNet~\cite{qi2017pointnet,qi2017pointnet++} pioneered direct point cloud processing. Recent advances in efficient point cloud encoders for online 3D scene perception have focused on balancing computational efficiency with high-quality feature extraction, enabling real-time understanding on resource-constrained platforms. Lightweight architectures, including voxel-based networks with sparse convolutions (e.g., PointPWC-Net~\cite{wu2019pointpwc}), point-based models using local neighborhood aggregation (e.g., PointNet++ variants~\cite{qi2017pointnet++}), and hybrid designs with attention mechanisms, have demonstrated improved performance with reduced latency. Innovations like KPConv~\cite{thomas2019kpconv} and PointTransformer~\cite{zhao2021point}, alongside recent quantization-aware and CPU-optimized models, have further accelerated inference with minimal accuracy reduction. Additionally, self-supervised pretraining (e.g., Sonata~\cite{wu2025sonata}) and knowledge distillation techniques have been employed to enhance feature representation while minimizing model complexity, making these encoders well-suited for robotics, AR/VR, and autonomous systems. \textit{Our work introduces an efficient 3D sparse feature pyramid network for this purpose.}

\paragraph{Online 3D Scene Perception.}
Online 3D scene perception, which involves precisely understanding a surrounding 3D environment from continuous RGB-D video streams, forms the crucial visual foundation for robotic tasks, including real-world applications like robotic navigation and manipulation~\cite{chaplot2020object,zhang20233d,mousavian20196}. Early methods approached this by processing 2D images independently and projecting the resulting predictions onto 3D point clouds, followed by a fusion step to integrate information across frames~\cite{mccormac2017semanticfusion,narita2019panopticfusion}. However, the lack of geometric and temporal awareness in 2D image predictions often led to complex and inaccurate fusion. To address these limitations, Fusion-aware 3D-Conv~\cite{zhang2020fusion} and SVCNN~\cite{huang2021supervoxel} constructed data structures to retain information from previous frames and performed point-based 3D aggregation for semantic segmentation. INS-CONV~\cite{liu2022ins} extends sparse convolution~\cite{graham20183d,choy20194d} to incremental CNNs, enabling efficient extraction of global 3D features for both semantic and instance segmentation. Aiming to simplify online 3D perception model design and leverage the capabilities of advanced offline 3D architectures, MemAda~\cite{xu2024memory} introduced a new paradigm using multimodal memory-based adapters to enable offline models with online perception abilities. In contrast to these prior works, ESAM~\cite{xu2024embodiedsam} lifts SAM-generated 2D masks to accurate 3D mask, facilitating efficient and highly accurate merging of per-frame predictions. However, ESAM's slow 3D sparse UNet hinders edge device deployment. \textit{In contrast, our method proposes to use an efficient 3D Sparse Feature Pyramid Network (SFPN) for fast multi-scale geometric feature extraction from streaming point clouds, significantly reducing computation and model size.}

\paragraph{2D Vision Foundation Models for 3D Scene Perception.}
Vision foundation models (VFMs)~\cite{oquab2023dinov2,kirillov2023segment,li2024segment} have revolutionized 2D computer vision, achieving remarkable accuracy and generalization thanks to large annotated datasets. Their success in zero-shot learning makes them attractive for 3D scene understanding, a field hampered by limited labeled data. Consequently, researchers are exploring how 2D VFMs can enhance 3D scene perception~\cite{rozenberszki2024unscene3d,yang2023sam3d,yin2024sai3d,xu2024embodiedsam}. For example, UnScene3D~\cite{rozenberszki2024unscene3d} uses self-supervised 2D features from DINO~\cite{oquab2023dinov2} to generate and refine pseudo 3D masks. SAM3D~\cite{yang2023sam3d} projects 2D instance masks, obtained from SAM~\cite{kirillov2023segment}, into 3D space using depth and camera information, then merges them based on geometry. SAMPro3D~\cite{xu2023sampro3d} treats 3D points as prompts for a multi-view 2D segmentation network, aligning the resulting masks and clustering 3D points. SAI3D~\cite{yin2024sai3d} generates 3D primitives and uses semantic-SAM~\cite{li2024segment} to obtain 2D masks with semantic scores, connecting and merging them using a graph-based approach. More recent works~\cite{qin2024langsplat,ye2024gaussian} segment 3D scenes represented by 3D Gaussians using VFMs, offering a novel perspective on 3D instance segmentation. ESAM~\cite{xu2024embodiedsam} also employs SAM for 3D instance segmentation but introduces learnable and online 2D-to-3D projection and 3D mask merging, enabling more accurate and real-time predictions. \emph{Our method builds upon this trend by utilizing an efficient 3D sparse feature pyramid network, specifically tailored for edge devices without GPU acceleration.}

\section{Method}
\label{sec:method}

We propose ESAM++, a lightweight and scalable online 3D scene perception framework that incorporates a 3D sparse feature pyramid (SFPN) for efficient point cloud feature extraction. In this section, we first define the task of online 3D scene perception (Section~\ref{sec:online_3d_scene_perception}). We then provide an overview of our proposed method (Section~\ref{sec:esam++_overview}), followed by a detailed introduction of 3D sparse feature pyramids (SFPN) for efficient point cloud representation (Section~\ref{sec:sparse_3d_feature_pyramid_networks}). Finally, we describe the 3D instance segmentation module (Section~\ref{sec:3d_point_cloud_segmentation}) and implementation details (Section~\ref{sec:implementation_details}).

\subsection{Problem Formulation}
\label{sec:online_3d_scene_perception}

\noindent\textbf{Notations.} Given a continuous stream of RGB-D frames denoted as $\mathcal{X}_t = \{x_1, x_2, ..., x_t\}$, where each frame $x_i = (I_i, D_i, T_i)$ consists of a color image $I_i \in \mathbb{R}^{H \times W \times 3}$, a depth map $D_i \in \mathbb{R}^{H \times W}$, and a camera pose $T_i \in \mathbb{R}^{4\times4}$, point clouds are generated to enable 3D scene reconstruction. Each point cloud $P_i \in \mathbb{R}^{N_i \times 3}$ is generated by projecting the depth map $D_i$ into 3D space using the corresponding camera pose $T_i$. The 3D scene is then incrementally reconstructed by aggregating the point clouds as $S_t = \bigcup_{i=1}^t P_i$. 

\noindent\textbf{Objective.} The objective of online 3D scene perception is to predict instance-level 3D segmentation masks for $S_t$, assigning a label $y_j \in \{1, \dots, K\}$ to each point $p_j \in S_t$. Here $K$ is the number of pre-defined categories. At time $t$, we only predict the instance masks $M_t^{cur}$ of current frame $P_t$. Then we merge $M_t^{cur}$ with the previously \emph{accumulated} instance masks $M_{t-1}^{acc}$ of $S_{t-1}$ and get the updated instance masks $M_{t}^{acc}$ of $S_t$. For the first frame, the previously accumulated instance mask is empty. The challenge of online 3D scene perception lies in maintaining an up-to-date segmentation of the 3D scene $S_t$ at every time step, enabling dynamic and incremental scene understanding as new frames arrive.

\begin{figure*}
  \centering
  \includegraphics[width=\linewidth]{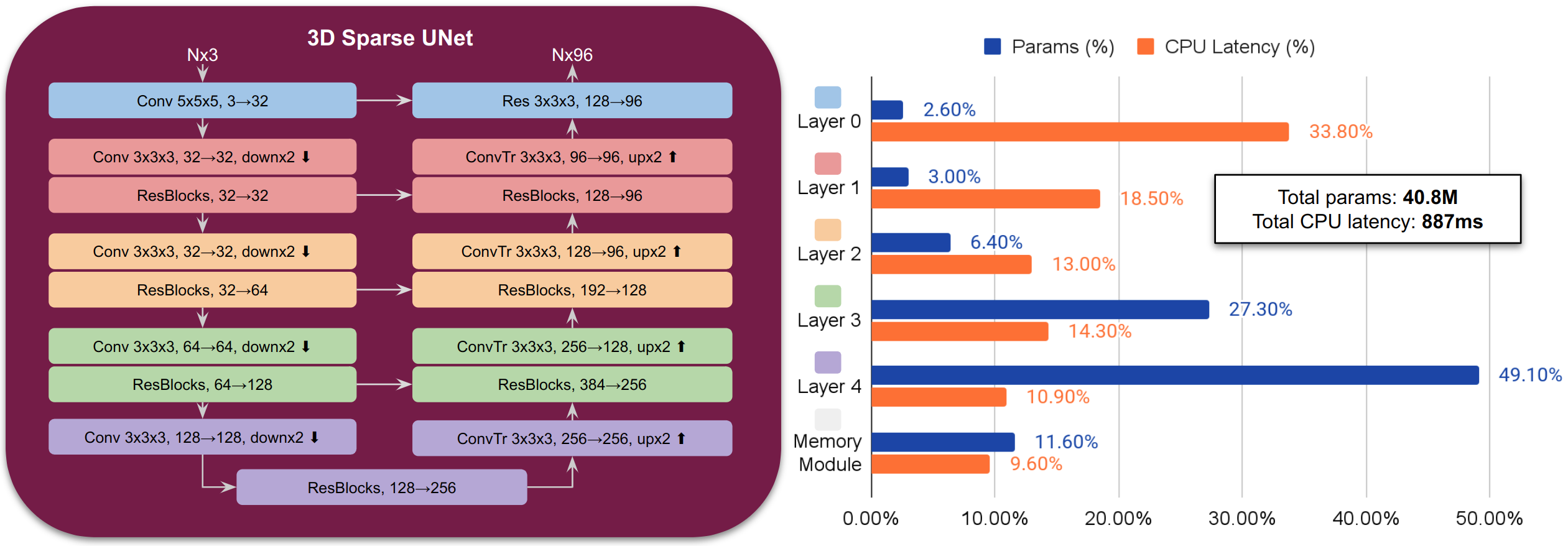}
  \caption{\textbf{Computational analysis of the 3D sparse UNet used in ESAM}. The left diagram shows architecture details; the right chart highlights parameter and latency distribution across each layer. Top layers (e.g., Layer 0) cause high latency due to voxel density and large kernels, while bottom layers (e.g., Layer 4) dominate model size. This motivates a more balanced encoder for edge use. The memory adapter module~\cite{xu2024memory} used in ESAM is not shown for brevity.}
  \label{fig:computational_analysis_for_3D_sparse_unet}
\end{figure*}

\subsection{ESAM++ Overview}
\label{sec:esam++_overview}

ESAM++ builds upon the architecture of ESAM~\cite{xu2024embodiedsam}, preserving its strength in online 3D instance segmentation while introducing a more efficient backbone for point cloud processing. As illustrated in Figure~\ref{fig:esam_framework_with_efficiency}, both frameworks adopt an incremental processing pipeline tailored for real-time scene understanding. However, whereas ESAM relies on a computationally intensive 3D sparse UNet~\cite{choy20194d} for point cloud feature extraction, ESAM++ replaces it with a lightweight and scalable 3D sparse feature pyramid (SFPN) designed to significantly reduce inference latency and model size.

\begin{figure*}
  \centering
  \includegraphics[width=\linewidth]{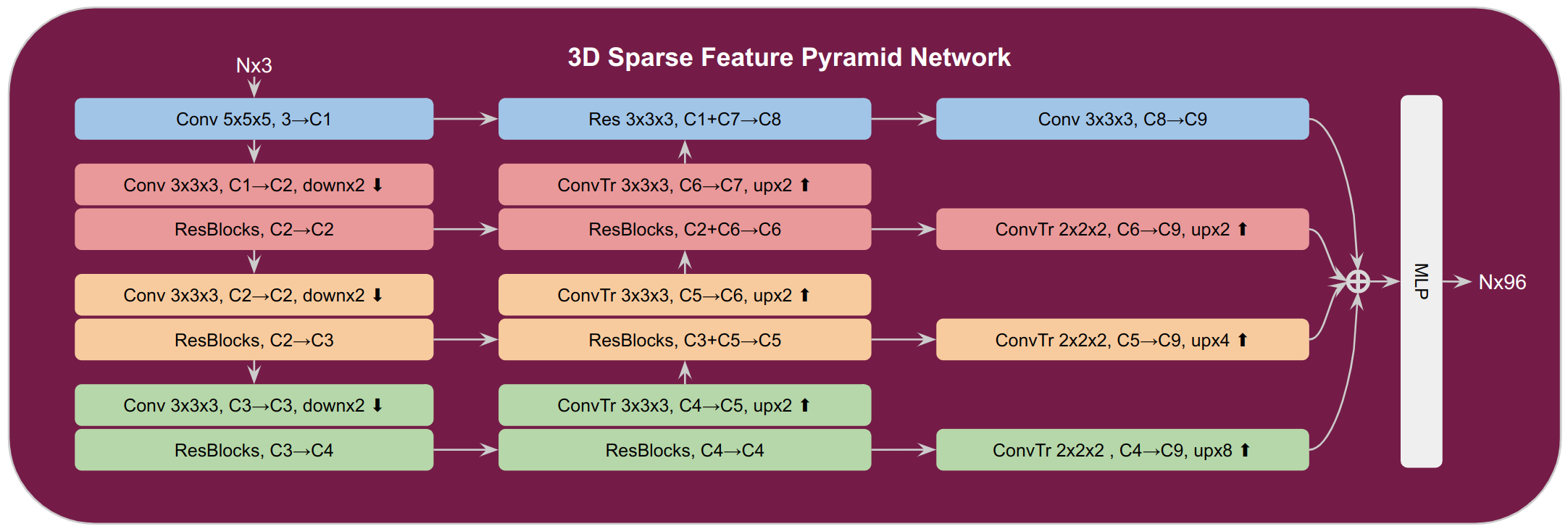}
  \caption{\textbf{Architecture of the proposed SFPN}. SFPN is a lightweight encoder-decoder for efficient multi-scale feature extraction from 3D point clouds. The encoder downsamples features through sparse convolutions and residual blocks, while the decoder upsamples and refines them. SFPN uniquely concatenates upsampled features from all decoder stages before an MLP generates the final point-wise features.  We implement three SFPN variants, with configurations summarized in Table~\ref{tab:sfpn_params}.}
  \label{fig:sfpn_architecture}
\end{figure*}

To motivate our design, we conduct a detailed computational analysis of the 3D sparse UNet in Figure~\ref{fig:computational_analysis_for_3D_sparse_unet}, identifying two primary inefficiencies: high-resolution (top) layers dominate the computation cost, while low-resolution (bottom) layers account for the majority of model parameters. Additionally, the decoder’s multi-scale feature pyramid is underutilized for prediction. Guided by these insights, we introduce SFPN (Figure~\ref{fig:sfpn_architecture}), a lightweight and scalable architecture that leverages multi-scale feature aggregation to preserve segmentation accuracy while significantly reducing both latency and model size—making it well-suited for deployment in resource-constrained environments. 

We now proceed to detail the key components of ESAM++, beginning with the SFPN architecture for sparse 3D feature extraction (Section~\ref{sec:sparse_3d_feature_pyramid_networks}), followed by the instance segmentation module (Section~\ref{sec:3d_point_cloud_segmentation}), and concluding with implementation details (Section~\ref{sec:implementation_details}).

\begin{table*}[]
    \centering
    \small
    \caption{Model parameters for SFPN variants.}
    \begin{tabular}{lcccccccccccc}
        \toprule
        \multirow{2}{*}{\textbf{Backbone}} & \multicolumn{9}{c}{\textbf{Channel Numbers}} & \multirow{2}{*}{\textbf{Model}} & \multirow{2}{*}{\textbf{CPU}} \\
        \cmidrule(lr){2-10}
        & C1 & C2 & C3 & C4 & C5 & C6 & C7 & C8 & C9 & \textbf{Params} & \textbf{Latency} \\
        \midrule
        SFPN - Small & 8 & 24 & 48 & 96 & 96 & 48 & 24 & 24 & 24 & 14.1M & 211ms \\
        SFPN - Base & 12 & 32 & 64 & 128 & 128 & 96 & 36 & 24 & 24 & 23.5M & 252ms \\
        SFPN - Large & 16 & 36 & 72 & 156 & 156 & 128 & 64 & 24 & 24 & 41.2M & 326ms \\
        \bottomrule
    \end{tabular}
    \label{tab:sfpn_params}
\end{table*}

\subsection{3D Sparse Feature Pyramid Network}
\label{sec:sparse_3d_feature_pyramid_networks}

The 3D Sparse Feature Pyramid Network (SFPN), illustrated in Figure~\ref{fig:sfpn_architecture}, features a lightweight encoder-decoder architecture designed for efficient multi-scale feature extraction from sparse 3D point clouds of shape $\mathbb{R}^{N\times3}$, outputting point cloud features $F_p \in \mathbb{R}^{N\times C}$, where $C$ is set to 96 by default. The encoder gradually downsamples the input through a series of sparse 3D convolutions and residual blocks, capturing hierarchical features across four spatial resolutions. Each stage expands the receptive field and reduces spatial resolution, with feature channels increasing from $C1$ to $C4$. The decoder mirrors this structure using sparse transposed convolutions (ConvTr) to progressively upsample features while refining them with residual blocks, with feature channels decreasing from $C4$ to $C8$. 

Unlike the 3D sparse UNet, which predicts solely from the topmost layer, SFPN harnesses the decoder’s multi-scale outputs by upsampling each to the original resolution via sparse transposed convolutions (ConvTr), followed by concatenation into a unified high-resolution feature map. This aggregated feature pyramid is then processed by a multi-layer perceptron (MLP) to produce the final point-wise features of dimension $C=96$, enabling accurate and efficient 3D segmentation with significantly reduced computational overhead. We implement three SFPN variants, with their channel configurations, model sizes, and CPU latencies summarized in Table~\ref{tab:sfpn_params}.

\subsection{3D Point Cloud Segmentation}
\label{sec:3d_point_cloud_segmentation}

This module predicts instance masks $M_t^{cur} \in \mathbb{R}^{N\times K}$ for the current frame, given the fine-grained 2D masks $M_t^{2d} \in \mathbb{R}^{H\times W \times L}$ produced by SAM and the point cloud features $F_p \in \mathbb{R}^{N\times C}$ extracted by our SFPN. Here, $L$ denotes the number of class-agnostic binary masks predicted by SAM, and $K$ represents the number of predefined semantic categories. Following the approach of ESAM~\cite{xu2024embodiedsam}, we achieve this via the following steps.

\noindent\textbf{Query lifting and refinement.} We lift $L$ 2D masks into a set of $L$ superpoints, denoted as $F_s \in \mathbb{R}^{L\times C}$, by pooling features from $F_p \in \mathbb{R}^{N\times C}$ according to the spatial support of each 2D mask. We initialize a set of 3D instance queries $Q_0$ from the lifted superpoint features $F_s$, which are then iteratively refined through a series of transformer-based query decoder layers and used to predict 3D instance masks $M_t^{cur} \in \mathbb{R}^{N\times K}$. 

\noindent\textbf{Query merging.} After lifting the 2D masks $M_t^{2d}$ to accurate 3D masks $M_t^{cur}$, we merge them with the previously accumulated instance masks $M_{t-1}^{acc}$. Thanks to the fixed-size feature representation for each mask in both $M_t^{cur}$ and $M_{t-1}^{acc}$, this merging process can be performed efficiently by computing pairwise similarities between the masks---implemented as a simple matrix multiplication.

\subsection{Implementation Details}
\label{sec:implementation_details}

For fair comparisons, we largely follow the experimental settings of ESAM~\cite{xu2024embodiedsam}, including the loss function and training strategies, which are included in the appendix. All experiments are conducted using PyTorch~\cite{paszke2019pytorch} on an NVIDIA A6000 GPU for training and Intel® Xeon® Silver 4314 CPUs (2.40 GHz) for inference. For more implementation details, please refer to the appendix.

\section{Experiments}

\subsection{Benchmarks}

We evaluate our method on both \emph{class-agnostic} and \emph{class-aware} 3D instance segmentation tasks across four datasets: ScanNet~\cite{dai2017scannet}, ScanNet200~\cite{rozenberszki2022language}, SceneNN~\cite{hua2016scenenn} and 3RScan~\cite{wald2019rio}. For class-agnostic segmentation, we train models on ScanNet200; for class-aware segmentation, we train models on ScanNet. Details are provided below. 

\paragraph{Datasets.} We evaluate on the following four datasets: 1) \textbf{ScanNet}, a large-scale indoor RGB-D dataset with 1,513 scenes and 3D semantic and instance annotations; we use 1,201 scenes for training and 312 for testing. 2) \textbf{ScanNet200}, an extension of  ScanNet with more fine-grained annotations, providing instance-level labels for over 200 semantic categories. 3) \textbf{SceneNN}, a 3D indoor scene dataset consisting of 50 high-quality RGB-D scans with detailed instance-level and semantic annotations. Following previous works~\cite{xu2024memory,xu2024embodiedsam}, we select 12 clean sequences for testing. 4) \textbf{3RScan}, a challenging indoor RGB-D dataset featuring over 1400 indoor scenes captured with fast camera motion, providing posed sequences and 3D reconstructions with semantic and instance-level annotation. We use its official test split, which includes 46 scenes, for evaluation.

\paragraph{Evaluation metrics.} We report Average Precision (AP) as the primary segmentation metric, calculated as the mean precision averaged across multiple Intersection over Union (IoU) thresholds---typically from 0.5 to 0.95 in steps of 0.05. In addition, we report AP$_{25}$ and AP$_{50}$, which correspond to average precision at fixed IoU thresholds of 0.25 and 0.50, respectively. We also report CPU latency (ms), measuring the average inference time per frame on CPU.

\begin{table*}
    \centering
    \small
    \caption{Class-agnostic 3D instance segmentation results of various methods on the ScanNet200 dataset.}
    \begin{tabular}{lccccccc}
        \toprule
        \multirow{2}{*}{\textbf{Method}} & \multirow{2}{*}{\textbf{Online}} & \multirow{2}{*}{\textbf{VFM}}
        & \multicolumn{3}{c}{\textbf{ScanNet200}} 
        & \multirow{2}{*}{\textbf{Model}} & \multirow{2}{*}{\textbf{CPU}} \\
        \cmidrule(lr){4-6}
        & & & AP & AP$_{50}$ & AP$_{25}$ & \textbf{Params} & \textbf{Latency} \\
        \midrule
        SAMPro3D~\cite{xu2023sampro3d}~\tiny{\emph{3DV'25}} & \xmark & SAM 
        & 18.0 & 32.8 & 56.1 & N/A & N/A \\
        Open3DIS~\cite{nguyen2024open3dis}~\tiny{\emph{CVPR'24}} & \xmark & GroundedSAM 
        & 34.6 & 43.1 & 48.5 & N/A & N/A \\
        SAI3D~\cite{yin2024sai3d}~\tiny{\emph{CVPR'24}} & \xmark & SemanticSAM 
        & 28.2 & 47.2 & 67.9 & N/A & N/A \\
        \midrule
        SAM3D~\cite{yang2023sam3d}~\tiny{\emph{ICCVW'23}} & \cmark & SAM 
        & 20.2 & 35.7 & 55.5 & N/A & T{\tiny{VFM}} + 18s \\
        ESAM~\cite{xu2024embodiedsam}~\tiny{\emph{ICLR'25}} & \cmark & SAM 
        & 42.2 & 63.7 & 79.6 & 44.6M & T{\tiny{VFM}} + 934ms \\
        ESAM-E~\cite{xu2024embodiedsam}~\tiny{\emph{ICLR'25}} & \cmark & FastSAM 
        & 43.4 & 65.4 & 80.9 & 44.6M & T{\tiny{VFM}} + 934ms \\
        \rowcolor{SkyBlue}Ours-Small & \cmark & FastSAM 
        & 30.3 & 55.8 & 68.9 & \textbf{14.1M} & T{\tiny{VFM}} + \textbf{211}ms \\
        \rowcolor{SkyBlue}Ours-Base & \cmark & FastSAM 
        & 39.7 & 60.6 & 77.5 & 23.5M & T{\tiny{VFM}} + 252ms \\
        \rowcolor{SkyBlue}Ours-Large & \cmark & FastSAM 
        & \textbf{43.7} & \textbf{66.1} & \textbf{81.2} & 41.2M & T{\tiny{VFM}} + 326ms \\
        \bottomrule
    \end{tabular}
    \label{tab:class_agnostic_segmentation}
\end{table*}

\begin{table*}
    \centering
    \small
    \caption{Cross-dataset generalization results from ScanNet200 to SceneNN and ScanNet200 to 3RScan. We directly evaluate the models from Table~\ref{tab:class_agnostic_segmentation} on these datasets to assess their transferability.}
    \begin{tabular}{lcccccccc}
        \toprule
        \multirow{2}{*}{\textbf{Method}} & \multirow{2}{*}{\textbf{Online}} & \multirow{2}{*}{\textbf{VFM}} 
        & \multicolumn{3}{c}{\textbf{ScanNet200}$\rightarrow$\textbf{SceneNN}} 
        & \multicolumn{3}{c}{\textbf{ScanNet200}$\rightarrow$\textbf{3RScan}} \\
        \cmidrule(lr){4-6} \cmidrule(lr){7-9}
        & & & AP & AP$_{50}$ & AP$_{25}$ & AP & AP$_{50}$ & AP$_{25}$ \\
        \midrule
        SAMPro3D~\cite{xu2023sampro3d}~\tiny{\emph{3DV'25}} & \xmark & SAM & 12.6 & 25.8 & 53.2 & 3.9 & 8.0 & 21.0  \\
        Open3DIS~\cite{nguyen2024open3dis}~\tiny{\emph{CVPR'24}} & \xmark & GroundedSAM & 18.2 & 32.2 & 48.9 & 9.5 & 21.8 & 47.0  \\
        SAI3D~\cite{yin2024sai3d}~\tiny{\emph{CVPR'24}}   & \xmark & SemanicSAM & 18.6 & 34.7 & 65.7 & 8.1 & 16.9 & 37.0 \\
        \midrule
        SAM3D~\cite{yang2023sam3d}~\tiny{\emph{ICCVW'23}}    & \cmark & SAM & 15.1 & 30.0 & 51.8 & 6.2 & 13.0 & 33.9 \\
        ESAM~\cite{xu2024embodiedsam}~\tiny{\emph{ICLR'25}}     & \cmark & SAM & 28.8 & 52.2 & 69.3 & 14.1 & 31.2 & 59.6 \\
        ESAM-E~\cite{xu2024embodiedsam}~\tiny{\emph{ICLR'25}}   & \cmark & FastSAM & 28.6 & 50.4 & 71.0 & 13.9 & 29.4 & 58.8 \\
        \rowcolor{SkyBlue}Ours-Small & \cmark & FastSAM 
        & 20.3 & 38.4 & 55.8 
        & 7.9 & 15.6 & 38.0 \\
        \rowcolor{SkyBlue}Ours-Base & \cmark & FastSAM 
        & 25.7 & 47.2 & 65.6 
        & 12.5 & 27.3 & 58.4 \\
        \rowcolor{SkyBlue}Ours-Large & \cmark & FastSAM 
        & \textbf{29.3} & \textbf{52.5} & \textbf{71.9} 
        & \textbf{14.4} & \textbf{31.7} & \textbf{60.3} \\        
        \bottomrule
    \end{tabular}
    \label{tab:cross_data_segmentation}
\end{table*}

\begin{table*}
    \centering
    \small
    \caption{Class-aware 3D instance segmentation results of different methods on ScanNet and SceneNN datasets. Our large model achieves state-of-the-art performance compared with previous online 3D instance segmentation methods.}
    \begin{tabular}{l@{\hspace{2pt}}cccccccc}
        \toprule
        \multirow{2}{*}{\textbf{Method}} & \multirow{2}{*}{\textbf{Online}} & \multirow{2}{*}{\textbf{VFM}} 
        & \multicolumn{3}{c}{\textbf{ScanNet}} 
        & \multicolumn{3}{c}{\textbf{SceneNN}} \\
        \cmidrule(lr){4-6} \cmidrule(lr){7-9}
        & & & AP & AP$_{50}$ & AP$_{25}$ & AP & AP$_{50}$ & AP$_{25}$ \\
        \midrule
        TD3D~\cite{kolodiazhnyi2024top}~\tiny{\emph{WACV'24}} & \xmark & N/A & 46.2 & 71.1 & 81.3 & - & - & -  \\
        Oneformer3D~\cite{kolodiazhnyi2024oneformer3d}~\tiny{\emph{CVPR'24}} & \xmark & N/A & 59.3 & 78.8 & 86.7 & - & - & - \\
        \midrule
        INS-Conv~\cite{liu2022ins}~\tiny{\emph{CVPR'22}} & \cmark & SAM & - & 57.4 & - & - & - & - \\
        TD3D-MA~\cite{xu2024memory}~\tiny{\emph{CVPR'24}} & \cmark & SAM & 39.0 & 60.5 & 71.3 & 26.0 & 42.8 & 59.2 \\
        ESAM-E~\cite{xu2024embodiedsam}~\tiny{\emph{ICLR'25}} & \cmark & FastSAM & 41.6 & 60.1 & 75.6 & 27.5 & 48.7 & 64.6 \\
        ESAM-E+FF~\cite{xu2024embodiedsam}~\tiny{\emph{ICLR'25}}   & \cmark & FastSAM & 42.6 & 61.9 & 77.1 & 33.3 & 53.6 & 62.5 \\
        \rowcolor{SkyBlue}Ours-Small & \cmark & FastSAM 
        & 32.5 & 52.7 & 59.9 
        & 28.3 & 39.6 & 47.4 \\
        \rowcolor{SkyBlue}Ours-Base & \cmark & FastSAM 
        & 39.2 & 58.0 & 66.4 
        & 30.1 & 47.9 & 56.8 \\
        \rowcolor{SkyBlue}Ours-Large & \cmark & FastSAM 
        & \textbf{43.7} & \textbf{63.5} & \textbf{78.6} 
        & \textbf{34.1} & \textbf{53.9} & \textbf{62.7} \\
        \bottomrule
    \end{tabular}
    \label{tab:class_aware_segmentation}
\end{table*}

\subsection{Baselines}
\label{sec:baseline}

We compare ESAM++ with both \emph{online} and \emph{offline} methods. For offline methods, the input of each scene is a reconstructed point cloud derived from a posed RGB-D video, where predictions are made on the reconstructed point clouds. For online methods, the input is a streaming RGB-D video with known camera parameters, and predictions are made frame by frame in a causal manner, without access to future frames or the full scene reconstruction.

\paragraph{Online methods.} We compare with three online methods: SAM3D~\cite{yang2023sam3d}, ESAM~\cite{xu2024embodiedsam}, INS-Conv\cite{liu2022ins}, and TD3D-MA\cite{xu2024memory}. We evaluate SAM3D only on class-agnostic 3D instance segmentation. It relies on a hand-crafted merging strategy, which significantly slows down inference. INS-Conv and TD3D-MA are evaluated only on class-aware 3D instance segmentation. ESAM is evaluated on both class-agnostic and class-aware settings. We also include two variants of ESAM in our evaluation: ESAM-E, which uses FastSAM as the visual foundation model (VFM), and ESAM-E+FF, which additionally leverages the VFM's features for 3D segmentation. Across these methods, the VFMs include SAM~\cite{kirillov2023segment}, GroundedSAM~\cite{ren2024grounded}, SemanticSAM~\cite{li2024segment}, and FastSAM~\cite{zhao2023fast}.

\paragraph{Offline methods.} We compare with five offline methods: SAMPro3D~\cite{xu2023sampro3d}, Open3DIS~\cite{nguyen2024open3dis}, SAI3D~\cite{yin2024sai3d}, TD3D~\cite{kolodiazhnyi2024top}, and Oneformer3D~\cite{kolodiazhnyi2024oneformer3d}. To ensure fairness, we adopt the 2D version of Open3DIS, since its 3D variant operates directly on reconstructed point clouds, while others predict from RGB-D frames. 

\subsection{Comparisons}

\paragraph{Class-agnostic 3D instance segmentation.}
We compare with VFM-assisted and online 3D instance segmentation methods, as detailed in Section~\ref{sec:baseline}. T{\tiny{VFM}} is the latency of SAM models. Our method is presented in three versions: Ours-Small, Ours-Base, and Ours-Large, all utilizing FastSAM for real-time inference. As shown in Table \ref{tab:class_agnostic_segmentation}, on the class-agnostic 3D instance segmentation task, our method sets a new state-of-the-art, surpassing previous methods, including offline approaches. It is important to note that online methods face greater challenges than offline alternatives, as offline methods process complete reconstructed 3D scenes, while online methods must handle partial and noisy frames. Despite these challenges, our method not only achieves high accuracy but also outperforms previous methods in speed. Leveraging an efficient architecture and a fast merging strategy, our method processes each frame in a mere 326ms on the CPU. This is a substantial improvement over ESAM, whose reliance on a computationally inefficient 3D UNet results in processing times nearly three times longer per frame. Our method achieves real-time online 3D instance segmentation without GPU acceleration while maintaining superior accuracy compared to previous methods.

\paragraph{Cross-dataset generalization evaluation.} 
Regarding generalization, our method also shows excellent performance. As demonstrated in Table \ref{tab:cross_data_segmentation}, when transferred to other datasets, our method continues to achieve state-of-the-art accuracy, outperforming zero-shot methods. In contrast, offline methods underperform on the 3RScan dataset, as they heavily depend on clean, reconstructed 3D meshes with accurately aligned RGB frames. In 3RScan, however, the fast camera movement results in blurry RGB images and camera poses, hindering the performance of offline methods.

\paragraph{Class-aware 3D instance segmentation.}
As shown in Table~\ref{tab:class_aware_segmentation}, our method achieves state-of-the-art performance compared with previous online 3D instance segmentation methods. Different from previous methods that only fuse 2D features to 3D point clouds, our approach utilize both 2D features and 2D masks to better guide the learning of 3D representation.

\begin{table}[h]
  \centering
  \small
  \renewcommand\arraystretch{0.9}
  \caption{Edge inference performance. Latency and power metrics are averaged over 100 iterations of a standard 3D perception pass..}  
  \begin{tabular}{l c c c}
    \toprule
    \textbf{Device} & \textbf{Chip} & \textbf{CPU Latency} & \textbf{Power} \\
    \midrule
    % \rowcolor{SkyBlue}    
    iPhone 15 & A16 Bionic & 190ms & 4.5W \\
    \bottomrule
  \end{tabular}
  \label{tab:edge_eval}
\end{table}

\paragraph{Evaluation on edge devices.} 
To evaluate real-world edge performance, we conducted experiments on an iPhone 15 equipped with the A16 Bionic chip. We specifically target the balance between computational throughput and thermal stability, as 3D perception tasks are notoriously resource-intensive.  As shown in Table~\ref{tab:edge_eval}, our small model achieves an average CPU latency of 190ms per frame. While the A16 Bionic features a powerful Neural Engine, we report CPU-based latency to establish a conservative baseline for devices without specialized AI accelerators. 

\subsection{Ablations and Analysis}

\begin{figure*}
  \centering
  \begin{subfigure}[b]{0.48\textwidth}
    \includegraphics[width=\textwidth]{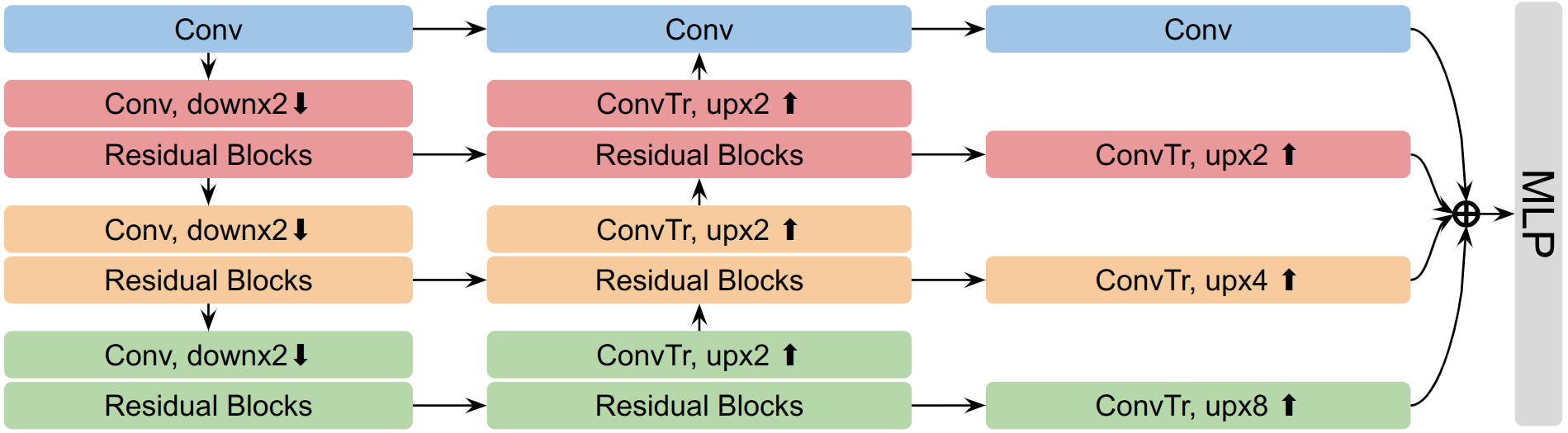}
    \caption{Full architecture.}
  \end{subfigure}
  \hfill
  \begin{subfigure}[b]{0.48\textwidth}
    \includegraphics[width=\textwidth]{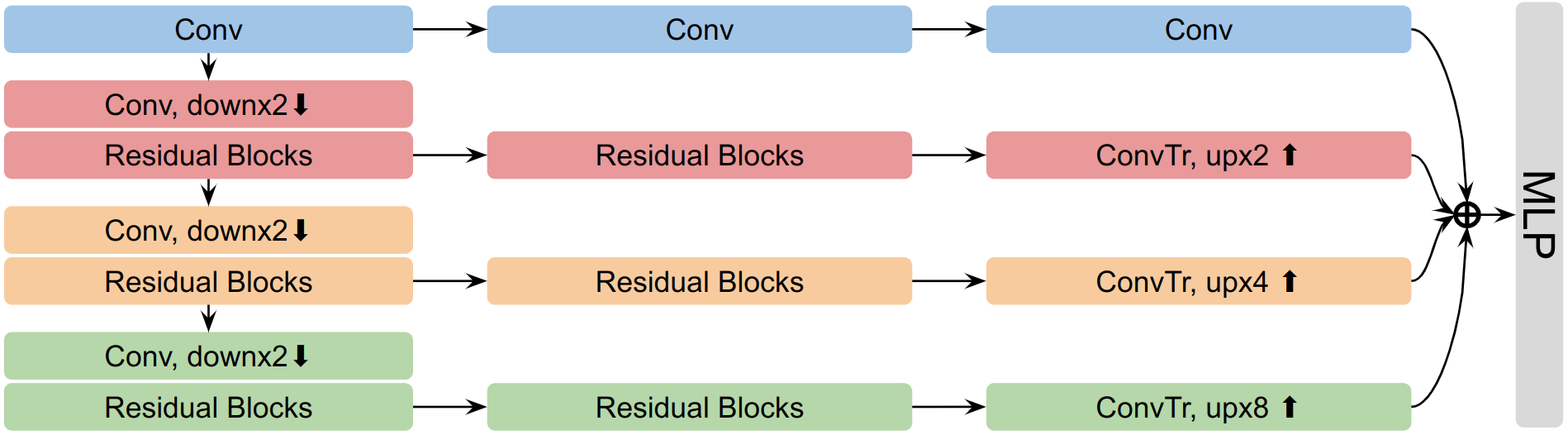}
    \caption{No upsampled fusion.}
  \end{subfigure}
  
  \vspace{0.2cm}  % Adjust as needed
  
  \begin{subfigure}[b]{0.48\textwidth}
    \includegraphics[width=\textwidth]{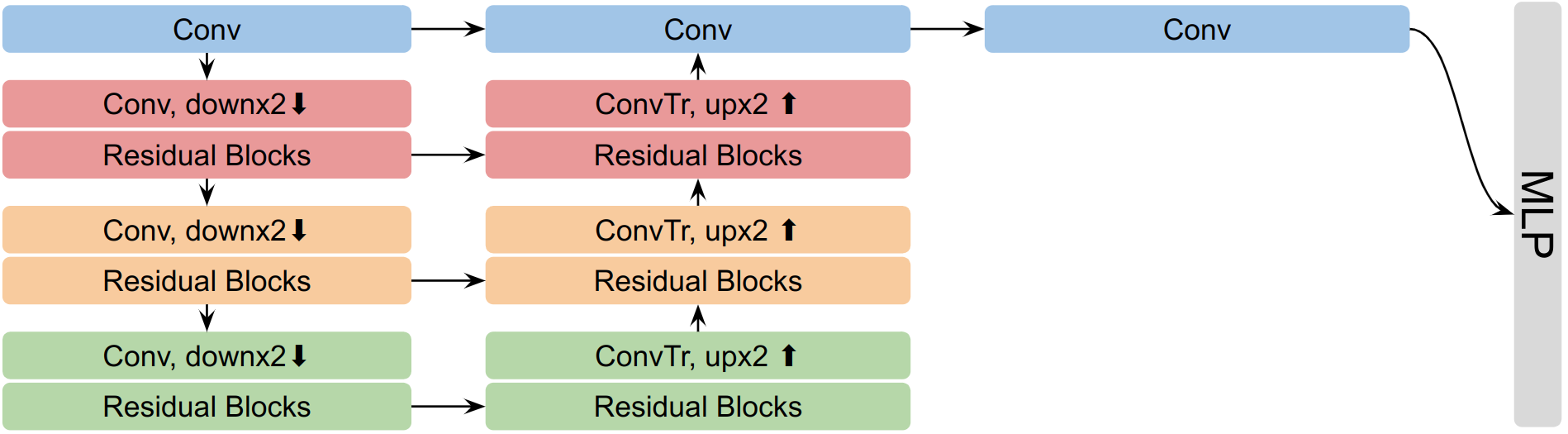}
    \caption{No pyramid.}
  \end{subfigure}
  \hfill
  \begin{subfigure}[b]{0.48\textwidth}
    \includegraphics[width=\textwidth]{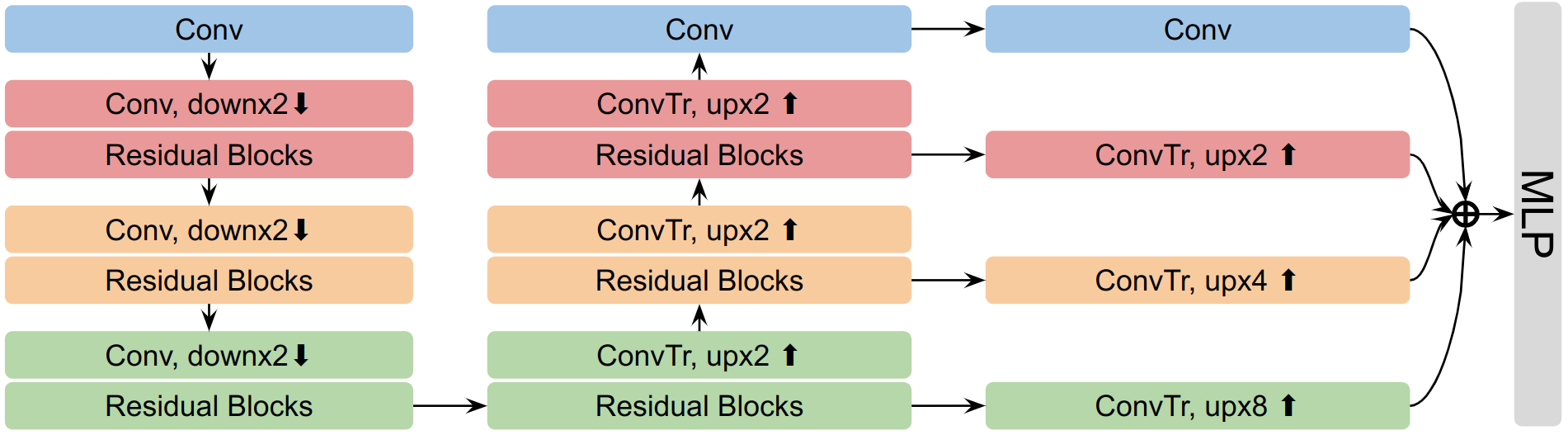}
    \caption{No skip connections.}
  \end{subfigure}
  \caption{\textbf{Ablation study} of the SFPN architecture: (a) full model, (b) without upsampled feature fusion, (c) without the feature pyramid, and (d) without skip connections. Comparisons results are shown in Table~\ref{tab:ablations_architecture}.}
  \label{fig:ablations_full_model}
\end{figure*}

\paragraph{Ablations.} We conduct ablation studies on ScanNet200 for class-agnostic 3D instance segmentation to validate the effectiveness of our SFPN design, as summarized in Table~\ref{tab:ablations_architecture} and visualized in Figure~\ref{fig:ablations_full_model}. Specifically, \emph{No upsampled fusion} disables the upsampling layers in the decoder, \emph{No pyramid} removes the hierarchical feature extraction, and \emph{No skip connection} eliminates skip connections between encoder and decoder. The results demonstrate the critical role for each component.

\begin{table}
\makeatletter\def\@captype{table}
\small
\centering
\caption{Effects of the architecture design.}
\begin{tabular}{lccc}
    \toprule
    Method & AP & Params & Latency \\
    \midrule
    No upsampled fusion & 34.0 & 39.6M & 312ms \\
    No pyramid & 33.8 & 39.6M & 298ms \\
    No skip connection & 36.4 & 36.5M & 277ms \\
    \rowcolor{SkyBlue}Full & 43.7 & 41.2M & 326ms \\
    \bottomrule
\end{tabular}
\label{tab:ablations_architecture}
\end{table}

\begin{table}
\makeatletter\def\@captype{table}
\small
\centering
\caption{Robustness to noisy camera poses.}
\begin{tabular}{cccc}
    \toprule
    Noise Ratio & AP & AP$_{50}$ & AP$_{25}$ \\
    \midrule
    \rowcolor{SkyBlue}0\% & 43.7 & 66.1 & 81.2 \\
    1\% & 43.7 & 66.0 & 81.2 \\
    5\% & 42.4 & 64.6 & 79.6 \\
    10\% & 38.9 & 60.3 & 77.4 \\
    \bottomrule
\end{tabular}
\label{tab:ablations_noise}
\end{table}

\begin{figure*}
  \centering
  \includegraphics[width=0.85\linewidth]{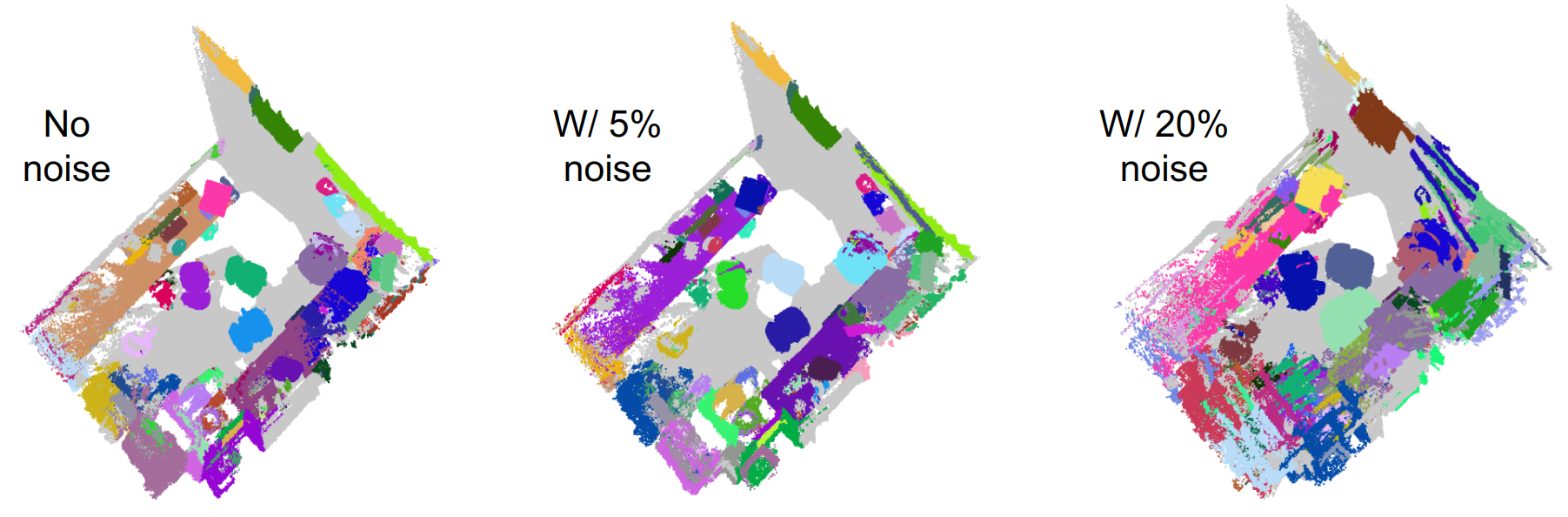}
  \caption{\textbf{Impact of noisy camera poses}. We evaluate our method on the ScanNet200 dataset for online class-agnostic 3D instance segmentation. The results show that our method remains robust under camera pose noise of up to 5\%. However, performance degrades significantly when the noise level increases to 20\%, leading to failure cases.}
  \label{fig:noise_comparison}
\end{figure*}

\paragraph{Impact of noisy camera poses.}
Both ESAM and ESAM++ require posed RGB-D videos with known camera intrinsics and extrinsics. To investigate the robustness of our method, we introduce noise to the camera poses by adding uniformly distributed random perturbations at varying scales. The quantitative results in Table~\ref{tab:ablations_noise} demonstrate the robustness of our method, showing only a modest performance drop even with 5\% pose noise. The qualitative results in Figure~\ref{fig:noise_comparison} illustrate that while performance noticeably degrades at higher noise levels (significant failures at 20\%), our method remains quite robust at a 5\% noise scale. These findings highlight the potential of our approach for online 3D scene perception using only RGB inputs by integrating with real-time reconstruction models like CUT3R~\cite{wang2025continuous}, which generate 3D scenes directly from RGB videos. We leave this promising direction for future exploration.
\section{Conclusion}
In this work, we presented ESAM++, a streamlined and efficient framework for online 3D scene perception in scenarios without GPU acceleration. Building upon ESAM, we replace its computationally expensive 3D sparse UNet with a novel lightweight Sparse Feature Pyramid Network (SFPN), enabling ESAM++ to dramatically reduce both inference latency and model size. Despite these improvements, ESAM++ retains competitive segmentation accuracy, achieving real-time performance on a CPU without sacrificing predictive quality. Through comprehensive evaluation on multiple benchmarks, we demonstrate that ESAM++ generalizes well across diverse environments, making it a practical solution for real-world applications on edge devices without GPU acceleration. 

\paragraph{Acknowledgment.} This project was partially supported by the Stanford HAI Hoffman-Yee Award. A portion of this work was conducted while the first author was a student researcher at Google. The authors would like to thank Zhenyang Shen and Lin Li for their insightful discussions.
{
    \small
    \bibliographystyle{ieeenat_fullname}
    \bibliography{main}
}


\end{document}

\maketitle
% \appendix

\section{Comparisons against other lightweight 3D backbones.} 
Most existing popular 3D backbones (e.g., PointTransformer~\cite{wu2024point} and Sonata~\cite{wu2025sonata}) are primarily optimized for GPU inference or offline processing, while our approach focuses on the specific challenges of \textbf{online}, \textbf{CPU-only} 3D perception. For a rigorous evaluation, we compare our method against a lightweight MinkowskiNet~\cite{choy20194d} variant and SPVNAS~\cite{tang2020searching}, a model specifically optimized for mobile-friendly and hardware-efficient performance. Results on ScanNet200 demonstrate that our method consistently outperforms all baselines.

\begin{table}[h]
  \centering
  \small
  \renewcommand\arraystretch{0.9}
  \begin{tabular}{l c c c c}
    \toprule
    \multirow{2}{*}{\textbf{Method}}
    & \multirow{2}{*}{\textbf{Backbone}}
    & \multirow{2}{*}{\textbf{Model}} 
    & \multirow{2}{*}{\textbf{CPU}} 
    & \multirow{2}{*}{\textbf{mAP}} \\
    \addlinespace[0.5em]    
    & & \textbf{Params} & \textbf{Latency} & \\
    \midrule
    ESAM-E & Sparse UNet & 44.6M & 934ms & 43.4 \\
    MinkNet18 & Sparse ResNet & 11.7M & 487ms & 27.1 \\
    SPVNAS & SPVCNN & 12.5M & 269ms & 22.8 \\
    \rowcolor{SkyBlue}    
    Ours & SFPN-Small & \textbf{14.1M} & \textbf{211ms} & 30.3 \\
    \rowcolor{SkyBlue}    
    Ours & SFPN-Base & 23.5M & 252ms & 39.7 \\
    \rowcolor{SkyBlue}
    Ours & SFPN-Large & 41.2M & 326ms & \textbf{43.7} \\
    \bottomrule
  \end{tabular}
  \caption{Comparisons with lightweight 3D backbones specifically optimized for CPU inference.}
  \label{tab:cutie_fair}
\end{table}

\section{Data-efficient learning.} 
Following the experimental setup of ESAM~\cite{xu2024embodiedsam}, we evaluate the class-agnostic performance of ESAM++ on ScanNet200 using reduced training sets (10\% and 50\%) as reported in Table~\ref{tab:ablations_data}. The results indicate a relatively minor performance degradation even with only half the training data, a trend consistent with the findings of ESAM.

\begin{table}
\small
\centering
\caption{Performance under varying training data proportions.}
\begin{tabular}{lccc}
    \toprule
    Data Proportion & AP & AP$_{50}$ & AP$_{25}$ \\
    \midrule
    \rowcolor{SkyBlue}100\% & 43.7 & 66.1 & 81.2 \\
    50\% & 40.6 & 64.4 & 80.1 \\
    10\% & 32.4 & 53.0 & 71.4 \\
    \bottomrule
\end{tabular}
\label{tab:ablations_data}
\end{table}

\section{Loss Function}
\label{sec:loss_function}
To effectively supervise the online 3D segmentation task, we use two complementary types of loss functions: per-frame loss and cross-frame loss. The \textbf{per-frame loss} supervises predictions independently at each time step by leveraging 2D SAM-based annotations lifted to 3D. In contrast, the \textbf{cross-frame loss} enforces temporal consistency across adjacent frames by introducing a contrastive objective that aligns instance-level features over time. We detail each component below.

\paragraph{Per-frame loss.}

Each RGB-D frame is annotated with consistent semantic and instance labels across time. Given these annotations, we compute per-frame losses based on the predictions from each query. Since queries $Q_t$ are lifted one-to-one from 2D SAM masks, we bypass complex label assignment and directly supervise each query using the corresponding 2D mask annotation. Assuming each 2D SAM mask corresponds to a single instance, we obtain the ground-truth semantic label and 2D instance mask for each query. These masks are projected to 3D using depth-based pixel correspondence, from which we derive the 3D instance mask and its axis-aligned bounding box. With the above annotations, we compute the binary classification loss $\mathcal{L}_{cls}^{t}$ with cross-entropy to distinguish foreground and background instances. The predicted 3D masks are supervised using binary cross-entropy \(\mathcal{L}_{\text{bce}}^t\) and Dice loss \(\mathcal{L}_{\text{dice}}^t\). Bounding box and semantic predictions are trained with IoU loss \(\mathcal{L}_{\text{iou}}^t\) and binary cross-entropy \(\mathcal{L}_{\text{sem}}^t\), respectively. The per-frame loss $\mathcal{L}_{1}$ is defined as follow:

\begin{equation}
    \mathcal{L}_{1}=\frac{1}{T}\sum_{t=1}^T(\alpha\mathcal{L}_{cls}^t+\mathcal{L}_{bce}^t+\mathcal{L}_{dice}^t+\beta\mathcal{L}_{iou}^t+\mathcal{L}_{sem}^t),
\end{equation}    
where we set both $\alpha$ and $\beta$ to 0.5 without tuning the parameters.

\paragraph{Cross-frame loss.} We formulate a contrastive loss across adjacent frames. This loss encourages feature consistency for the same instance across adjacent frames $f_t$ and $f_{t+1}$. The cross-frame loss $\mathcal{L}_{2}$ is defined as follow:

\begin{equation}
    {\mathcal{L}_{2}} = \frac{1}{T}\sum_{t=1}^T({\mathcal{L}_{cont}}^{t \to {t+1}} + {\mathcal{L}_{cont}}^{t \to {t-1}}).
\end{equation}

\begin{equation}
    {\mathcal{L}_{cont}}^{t \to {t+1}} =- \frac{1}{Z} {\sum_{i=1}^{Z} \log \frac{e^{\left(\left<{f}^i_t,{f}_{t+1}^i\right>/\tau\right)}}{\sum_{j \neq i} e^{\left(\left<{f}_t^i,{f}_{t+1}^j\right>/\tau\right)}+e^{\left(\left<{f}_t^i,{f}_{t+1}^i\right>/\tau\right)}}},
\end{equation}

where $\left<\cdot,\cdot\right>$ is cosine similarity; the corner cases ${\mathcal{L}_{cont}}^{T \to {T+1}}$ and ${\mathcal{L}_{cont}}^{1 \to {0}}$ are set to zero.

\paragraph{Total loss.}
The total loss is formulated as:
\begin{equation}
    \mathcal{L}_{total}=\lambda_{1}{\mathcal{L}_{1}} + \lambda_{2}{\mathcal{L}_{2}}
\end{equation}
where we set both $\lambda_1$ and $\lambda_2$ to 0.5 without tuning the parameters.

\section{Training Strategies}
Following prior work~\cite{xu2024embodiedsam,xu2024memory}, we adopt a two-stage training strategy for ESAM++. In the first stage, we train a single-view perception model using ScanNet(200)-25k, a curated subset of ScanNet(200) containing individual RGB-D frames. At this stage, we exclude memory-based adapters and auxiliary task losses to focus solely on learning robust single-frame representations. This enables the model to develop a strong foundational understanding of object semantics and geometry from isolated views without the added complexity of temporal or memory-based reasoning.

In the second stage, we fine-tune the pretrained model on full RGB-D sequences, incorporating the memory-based adapters and the complete set of loss functions associated with auxiliary tasks. This stage is designed to enhance the model’s ability to integrate information across frames and maintain consistent object representations over time. To ensure computational efficiency and manage memory usage during training, we randomly sample 8 consecutive RGB-D frames from each scene at every training iteration. This sampling strategy allows the model to learn from temporal context while keeping the memory footprint tractable.

All experiments are implemented using PyTorch~\cite{paszke2019pytorch}. Model training is conducted on an NVIDIA A6000 GPU, while inference is performed on Intel® Xeon® Silver 4314 CPUs running at 2.40 GHz, demonstrating the efficiency and deployability of our approach in resource-constrained environments. We use the AdamW optimizer~\cite{loshchilov2017decoupled} with an initial learning rate of 1e-4 and a weight decay of 0.05 to ensure stable optimization and effective regularization throughout both training stages.

\section{Qualitative Results}
We present three qualitative demos to illustrate the online 3D segmentation process, as shown in Figure~\ref{fig:demo_case1}, Figure~\ref{fig:demo_case2}, and Figure~\ref{fig:demo_case3}. These examples are randomly selected from the ScanNet200 datasets, demonstrating that ESAM++ effectively merges partial segmentation results into complete objects and produces fine-grained 3D masks within the online reconstructed scenes. Please refer to the accompanying demo videos for further details.

\section{Combining with Online 3D Reconstruction Models}

While our method currently relies on depth cameras for 3D reconstruction, it is both interesting and promising to explore its integration with online 3D reconstruction models~\cite{wang2025continuous} that operate solely on RGB inputs. Such a combination would enable the development of an online 3D perception system that functions using only RGB video, removing the dependency on depth sensors. We consider this an exciting direction for future work.

\begin{figure*}
  \centering
  \includegraphics[width=\linewidth]{figs/supp_case1.png}
  \caption{Online visualization---case study 1.}
  \label{fig:demo_case1}
\end{figure*}

\begin{figure*}
  \centering
  \includegraphics[width=\linewidth]{figs/supp_case2.png}
  \caption{Online visualization---case study 2.}
  \label{fig:demo_case2}
\end{figure*}

\begin{figure*}
  \centering
  \includegraphics[width=\linewidth]{figs/supp_case3.png}
  \caption{Online visualization---case study 3.}
  \label{fig:demo_case3}
\end{figure*}

{
    \small
    \bibliographystyle{ieeenat_fullname}
    \bibliography{supp}
}